%% file: main.tex
\documentclass[conference,10pt, a4paper]{ieeeconf}
\pdfoutput=1
\usepackage{url}
\usepackage{cite}
\usepackage{amsmath,amssymb,amsfonts}
\usepackage{graphicx}
\usepackage{textcomp}
\usepackage[linesnumbered,ruled,lined,boxed,commentsnumbered]{algorithm2e}
\usepackage{algorithmicx}
\usepackage[noend]{algpseudocode}

\usepackage{csquotes}
\usepackage{todonotes}
\usepackage{multirow}
\usepackage{siunitx}
\usepackage{flushend}
\usepackage{subcaption}
\usepackage[absolute,overlay]{textpos}

\IEEEoverridecommandlockouts         
\newcommand{\opstop}{\textit{stop}}
\newcommand{\oppush}{\textit{push}}
\newcommand{\oprotate}{\textit{rotate}}
\newcommand{\opreplan}{\textit{replan}}
\newcommand{\opswap}{\textit{swap}}

\let\oldnl\nl
\newcommand{\nonl}{\renewcommand{\nl}{\let\nl\oldnl}}

\newlength\lenKwIn

\newlength\lenKwOut

\begin{document}
\begin{textblock*}{\textwidth}(1in+\hoffset+\oddsidemargin,1cm) 
\centering
\small
In: 2019 IEEE Intelligent Transportation Systems Conference (ITSC). IEEE Xplore, 2019. p. 4456-4463. ISBN 978-1-5386-7024-8.
\end{textblock*}

\title{\LARGE\bf Push, Stop, and Replan: An Application of Pebble Motion on Graphs to Planning in Automated Warehouses}

\author{
Miroslav Kulich$^{1}$, Tom\'a\v{s} Nov\'ak$^{2}$, and Libor P\v{r}eu\v{c}il$^{1}$
\thanks{Miroslav Kulich and Libor P\v{r}eu\v{c}il are with Czech Institute of Informatics, Robotics, and Cybernetics,
Czech Technical University in Prague, Prague, Czech Republic {\tt\small \{kulich,preucil\}@cvut.cz}}
\thanks{Tom\'a\v{s} Nov\'ak$^{2}$ is with the Department of Cybernetics, Faculty of Electrical Engineering, 
Czech Technical University in Prague, Prague, Czech Republic {\tt\small tomasnovakmail@gmail.com}}
\thanks{This work has been supported by the European Union's Horizon 2020 research and innovation programme under grant agreement No 688117, by the Technology Agency of the Czech Republic under the project no.~TE01020197 \enquote{Centre for Applied Cybernetics}, and by the European Regional Development Fund under the project Robotics for Industry 4.0 (reg. no. CZ.02.1.01/0.0/0.0/15 003/0000470).}}

\maketitle
\thispagestyle{empty}
\pagestyle{empty}
\begin{abstract}
  The pebble-motion on graphs is a subcategory of multi-agent pathfinding problems dealing with moving multiple pebble-like objects from a node to a node in a graph with a constraint that only one pebble can occupy one node at a given time.
  Additionally, algorithms solving this problem assume that individual pebbles (robots) cannot move at the same time and their movement is discrete. 
  These assumptions disqualify them from being directly used in practical applications, although they have otherwise nice theoretical properties.
  We present modifications of the Push and Rotate algorithm~\cite{DeWilde2014}, which relax the presumptions mentioned above and demonstrate, through a set of experiments, that the modified algorithm is applicable for planning in automated warehouses.


\end{abstract}


\input{intro}
\input{problem}

\input{algorithm}

\section{conclusion}
\label{sec:conclusion}
A novel planning algorithm for coordination of a robotic fleet in an automated warehouse based on a pebble motion on a graph algorithm, Push and Rotate, is presented. 
The algorithm allows a continuous movement of robots on their trajectories instead of a discrete movement between nodes. 
It also takes into consideration the rotation of the robots and their different velocities. 
One of the main achievements is that all the robots can move in parallel, which was not possible with the original P\&R. 
This is achieved by an implemented system of priorities that assures that the robot with the longest trajectory always moves
towards its final destination. Whenever the robots are involved in a push operation, the accumulation of priorities overcomes problems with deadlocks.
The computational time for the solution that has to be calculated before the result can be executed was another challenge. 
The algorithm solves this issue inherently by taking the shortest trajectories to destinations. 
The trajectories are only modified during the calculation. 
Thus if the algorithm is faster than real-time, the solution can be executed before the calculation is finished. 
The algorithm also allows adding robots during the calculation.

In future work, we want to study both the theoretical and practical aspect of the algorithm more thoroughly. 
From the theoretical point of view, the aim will be to prove the completeness of the algorithm as well as to derive bounds for the buffering time.
Moreover, we want to perform experiments in more challenging setups, i.e., scenarios in larger maps and involving more robots.
We are also preparing experiments in a real environment. 



\bibliographystyle{IEEEtran}
\bibliography{IEEEabrv,main}

\end{document}

%% file: intro.tex
Warehouses are used by industries to store assembly parts or goods to be sold. 
These warehouses often already have a computer system that tracks the position of the product in the racks, but the goods are usually moved to and from the racks by human beings. 
They navigate through the space between the racks searching for the correct rack and then they search for the item. 
When assembling an order composed by several different items, they usually spend a lot of the time walking around the warehouse. 
Companies like Amazon or SwissLog have therefore implemented solutions for automated warehouses where robots bring whole racks to picking stations. 
Here the human workers pick up the desired items and pack them into boxes according to orders. 
Such a system requires the robots to be able to navigate in the warehouse and avoid obstacles which could be static -- such as walls and racks -- or dynamic, mainly other robots and occasionally human beings.

The problem of coordinating a fleet of robots is called multi-agent pathfinding, which is known to be NP-complete for a discrete graph and PSPACE-complete for real environments~\cite{Hopcroft1984,Goldreich11}.
Traditional planning methods can be categorized into centralized and decentralized.
Approaches from the first category systematically search a composite configuration space built as a Cartesian product of particular robots' configurations and provide optimal solutions~\cite{Latombe1991,Lavalle1998,Ryan2008}.
On the other hand, the thorough search is very time consuming and only small instances can be thus solved.
Although many authors developed advanced search space pruning, which decreases the time complexity~\cite{Berg2009,Geramifard2006biased,Peasgood08,Wang2008}, the computational complexity is still high: problems with tens of robots are solved in minutes.

Decoupled approaches plan paths for individual robots independently from each other, which is followed by coordination of the robots~\cite{LaValle1998_OMP,Simeon2002}.
Alternatively, prioritized planning is used, which computes trajectories sequentially for individual robots based on their priorities. 
Robots with already determined trajectories are considered as moving obstacles to be avoided by robots with lower priorities~\cite{VandenBerg2005,Bennewitz2001,Cap2015}.

Several computationally efficient heuristics have been introduced recently~\cite{Chiew2010,Wang2011,Silver2005}.
Furthermore, another stream of research is based on a solution of the problem called pebble motion on graphs, the planning problem where only one agent moves at a time (the $15$-puzzle is the most famous example of this problem).
Luna and  Bekris~\cite{Luna2011} present a complete heuristics for general problems with at most $n-2$ robots in a graph with $n$ vertices based on the combination of two primitives - \enquote{push} forces robots towards a specific path, while \enquote{swap} switches positions of two robots if they are to be colliding. 
An extension which divides the graph into subgraphs within which it is possible for agents to reach any position of the subgraph, and then uses \enquote{push}, \enquote{swap}, and \enquote{rotate} operations is presented in~\cite{DeWilde2014}.

The main shortcoming of the pebble-motion solving algorithms is that individual agents cannot move at the same time. 
Therefore, the real usage of such a solution in a warehouse would be time-wasting and ineffective. 
All of the abovementioned algorithms also assume one type of robots, while robots in an automated warehouse might be split into those with and those without racks and assume a constant time of node-to-node movements, which is not possible in a real application. 
In this paper, we propose a modification of the Push and Rotate (P\&R) algorithm~\cite{DeWilde2014}, which is designed for the use in real warehouse applications by allowing parallel movement of two types of robots (loaded and unloaded) and accounts for non-constant node-to-node movement time. 
The proposed algorithm moves robots on the shortest possible trajectories to their destinations and in case of a conflict it uses a modified \oppush{} operation or one of the newly proposed operations: \opstop{} and \opreplan{}.

The rest of the paper is organized as follows. 
First, the problem is described in Section~\ref{sec:problem}.
The planning algorithm is detailed in Section~\ref{sec:algorithm}, while experimental results are presented and discussed in Section~\ref{sec:experiments}.
Section~\ref{sec:conclusion} is finally dedicated to concluding remarks.

%% file: problem.tex
\section{Problem definition}
\label{sec:problem}

The complete solution for robot management in an automated warehouse is typically composed of three layers. 
The highest layer is the warehouse management system (WMS), which creates a queue of tasks that are supposed to be accomplished from an order that is currently processed. 
The order may consist of several items that are stored at different locations.
Their position is thus determined, and the manager adds the information which rack has to be brought to which picking station.  
The middle layer processes the queue from WMS and determines an appropriate robot for each task.  
The lowest layer, the path-planning algorithm, coordinates the movement of the robots by computing their collision-free trajectories given their start and goal positions.

The path planning in the last layer is addressed in this paper. 
Specifically, we assume a connected, directed graph $G=(N,E)$, where $N$ is a set of nodes and $E$ is a set of connections between them.
For each robot $r_i\in R$  start $s_i$ and goal $g_i$ nodes are given as well as a subset of edges $E$ where the robot is allowed to operate.  
The aim is to find a set of non-colliding trajectories from $s_i$ to $g_i$ for each robot while minimizing some global cost function. A typical cost function is a sum of robots' travel times or a plan completion time.

\begin{figure}
   \centering
   \includegraphics[width=\columnwidth]{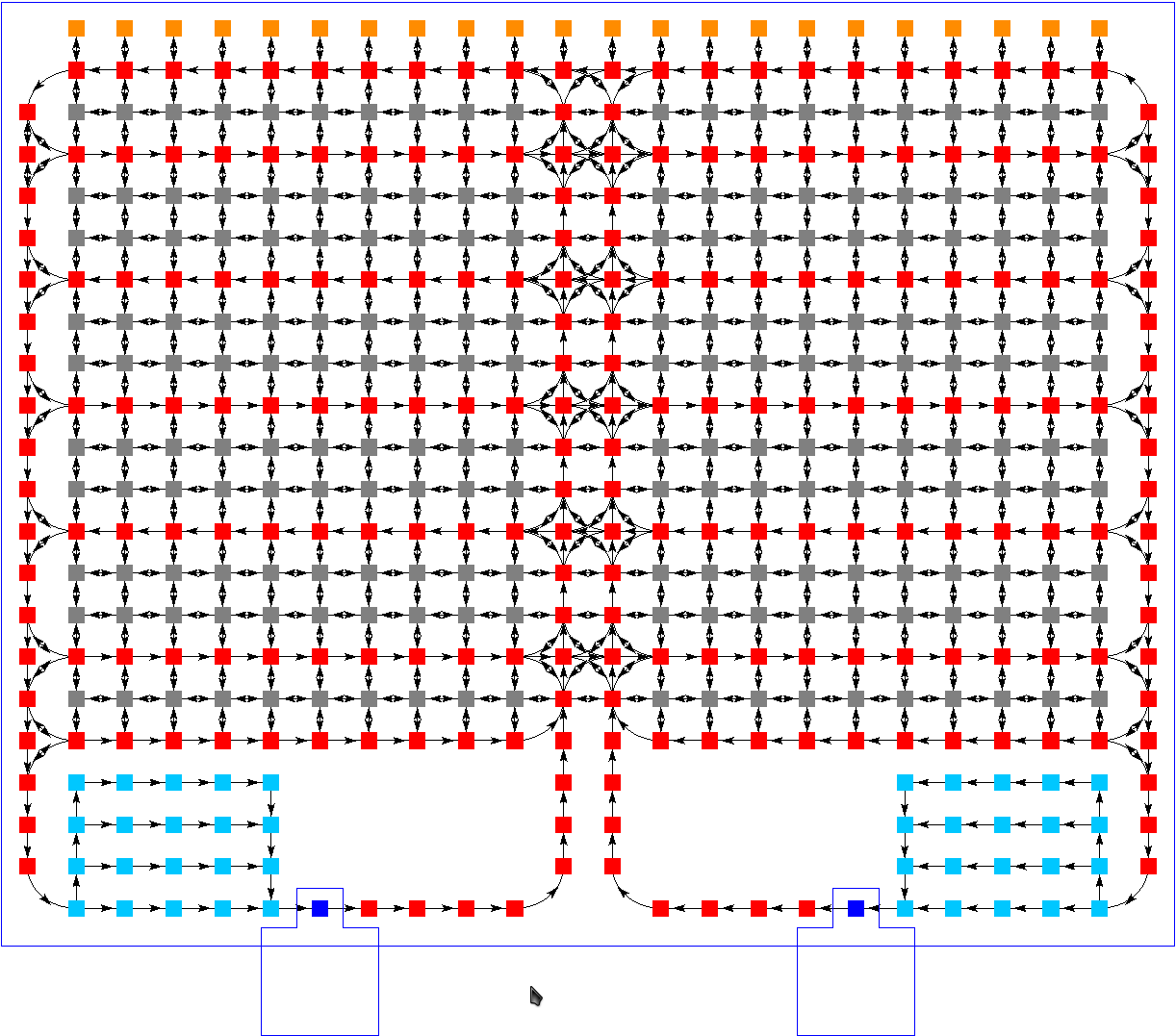}
   \caption{A map of a real warehouse.}
   \label{fig:real_graph}
\end{figure}

Fig.~\ref{fig:real_graph} illustrates a map of a typical warehouse: nodes on roads (red), positions of racks (grey), picking stations (blue polygons), maintenance nodes/charging stations (orange), queues before picking stations (light blue). Note that robots with loaded racks cannot go to grey nodes if they are occupied by some racks, while unloaded robots can.   
Two types of edges are distinguished: straight lines (we call them {\em default}) and turns, which allow robots to concurrently move and rotate {\em spline}. 

%% file: algorithm.tex
\section{Proposed algorithm}
\label{sec:algorithm}

\subsection{Warehouse-related requirements}
Several challenges appear when designing an algorithm for real continuous environments instead of for a discrete world of graphs. 
On the other hand, one can employ several simplifications in comparison to general graph algorithms using typical
properties of a warehouse to simplify the topology of the associated graph.
The main challenges that had to be resolved during the algorithm design are discussed in this section.


\subsubsection{Non-constant time of movement between nodes}
Contrary to standard graph algorithms, where robots move between nodes discretely, they can also occupy space between the nodes.
This movement is represented by a sequence of time steps containing necessary information about the robots: time, position, rotation, velocity and a set of occupied nodes. 
A model of a robot's movement is necessary to generate these sequences. 
The precision of this model defines the usability of the real scenario; however, in the proposed algorithm we only use a very simple model and propose a modification to deal with its inaccuracy in the real world scenario.

\subsubsection{Node/edge conflicts, mainly at spline edges and complicated junctions}
\begin{figure}
    \centering
    \begin{subfigure}[b]{0.3\columnwidth}
        \includegraphics[width=\textwidth]{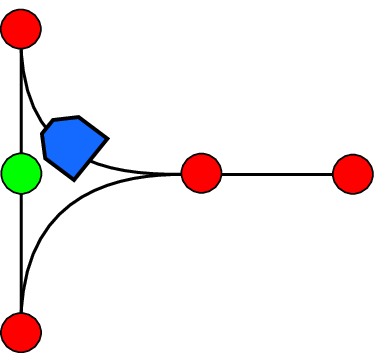}
        \caption{Spline shortcuts (the green node is in conflict with the robot).}
        \label{fig:junc_comlex1}
    \end{subfigure}
    \hfill
        \begin{subfigure}[b]{0.65\columnwidth}
            \raisebox{3.3em}{
        \includegraphics[width=\textwidth]{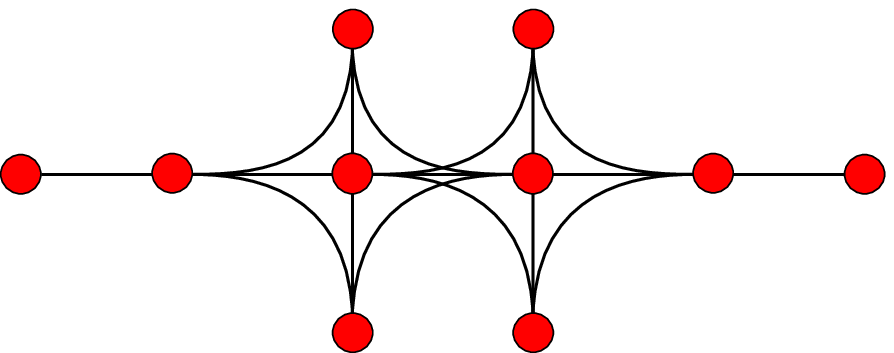}
            }
        \caption{Complex spline junctions}
        \label{fig:junc_comlex2}
    \end{subfigure}
    \caption{Complex spline junctions at the corners of aisles in the warehouse.}
\label{fig:junc_comlex}
\end{figure}

A set of conflicting nodes and edges that no robot can be present at the same time is defined for every node and edge. This is mainly the case of spline edges and complicated junctions as in Fig.~\ref{fig:junc_comlex}. 

\subsubsection{Parallel movement of robots} is necessary to reduce the makespan of the solution and make it viable for usage in a real warehouse. 
To achieve this goal, all robots are moved simultaneously at time-step. 
When one robot happens to conflict with another robot, one of the operations described in Section \ref{sec:alg_desc} is invoked to resolve the conflict. 
All of the operations involve only the robots in conflict and the positions of the task-accomplished robots. 
The consideration of a simultaneous movement of robots causes an increase in the complexity of the algorithm, but also adds more flexibility.

\subsubsection{Simplifications} A real warehouse environment increases the complexity of the algorithm significantly; however, the graph of the warehouse has properties that can be used to simplify the planning algorithm. 
It is assumed that the graph is always biconnected.
The decomposition in the P\&Rotate algorithm thus always ends up with one component, and, therefore, it can be omitted. 
Moreover, the \opswap{} operation always successfully completed as proved in \cite{DeWilde2014} and the \oprotate{} operation can be thus also omitted.

\subsubsection{Runtime requirements}
Robots in a warehouse are not given their tasks at once, but rather the goals are assigned one by one from the warehouse management system. 
When a goal is assigned to the robot, the time to find the solution should be minimized, which is a challenging requirement to meet. 
The algorithm should thus be able to reuse the already evaluated solution and add a new robot's movement as fast as possible to it.

\subsubsection{On-the-fly processing} 
The algorithm produces plans on the fly.
This means that first parts of plans are available and robots can start moving according to them before the planner finishes.
The only problem is in case of collision when the planner virtually moves the time backward to solve the collision.
In the extreme case when one collision resolution immediately causes a new conflict, the time can be moved backward significantly.
Therefore, some solution buffer has to be introduced guaranteeing that the virtual time in the planner is not behind the real-time.
To do that, a statistical evaluation to calculate the size of the solution buffer has been done. 
Nevertheless, it can happen that the buffer would get too small during the movement of the robots; the movement would have to be paused in this situation.

\subsection{Algorithm description}
\label{sec:alg_desc}
The proposed planning algorithm consists of three parts: the single robot path planning phase, the initial trajectory generation, and the robot maneuvering phase, and uses operations \opstop{}, \oppush{}, and \opreplan{} to modify the trajectories in case of conflict. 
Moreover, the current state and time of the warehouse is maintained. 

The priority of each robot is, initially, equivalent to length of the shortest paths from its origin to its destinations 
Besides this, a temporary priority of each robot is initialized to these constant priorities, incremented by the \oppush{} operation and reset to the value of the main priority. 
The \opstop{} operation tries to resolve the conflict by stopping one of the robots, while the \oppush{} operation pushes robot with a lower temporary priority out of the path of the other conflicting robot.

\subsubsection{Single robot path planning phase}
\label{sec:path_plan}


The shortest paths for all robots from their start to their goal positions are generated on the graph $G$ making use of the standard A* algorithm~\cite{astar}. 
The used heuristic function is the Euclidean distance to the goal, while the determination of edge cost depends on the edge type. We define the cost $C_l$ for spline edges and  $C_s$ for edges to storage location nodes bellow, while the default cost $C_d=1$ is used for all other edges. 

Traveling through spline edge is shorter than travel trough two default edges, but longer than traveling through one; thus $C_d < C_l < 2C_d$ and is set to $1.5$. 
The edges that end in the storage location nodes are in most cases traveled by the robots without racks. 
To enforce their preference to travel under the racks and leave more space on the road nodes for robots with racks, the cost $C_s$ must be $0<C_s < C_d$ and $0<C_s < \frac{C_I}{2}$, and is set to $0.1$. 
The robots that carry a rack cannot use edges starting or ending at the storage location nodes except initial and goal state of the robots.

\subsubsection{Initial trajectory generation phase}
The generated paths are processed by the robot model. 
The output for each robot $r_x$ is a list of time steps $J_r=\{j_{1},j_{2}, \ldots , j_{n}\}$, where $n$ is the number of time steps in the path. 
The output data depends on the model parameters. 
Because we need to discretize continuous movements, the time steps are samples of real trajectory movement; thus sample frequency must be defined reasonably. 
The algorithm directly processes the time samples.
Therefore the computational difficulty grows with the sampling frequency. 
Choosing a too small number of samples could lead to failure in case that there are not at least $2$ samples between two nodes -- each where a robot occupies one of the two nodes on the edge. 
The robots could come into a conflict in moments that were not captured by the samples; thus the algorithm would not detect it.
It is reasonable to have at least ten samples for each edge. 
To make sure the sampling frequency $F_s$ is sufficient, it should meet the condition $F_s > \frac{l_{min}}{v_{max}} \times 10$, where $l_{min}$ is the minimum length of an edge and $v_{max}$ is the maximal robot velocity. 

\begin{algorithm}[ht]
\DontPrintSemicolon
 \KwData{generated trajectories $J$}
  \KwResult{Trajectories $J$, modified to be collision-free.}
  $solved \leftarrow false$\;
 \While{$!solved$}{\label{alg:solver_solved}
  	\If{$conflict\_detect()$}{\label{alg:solver_conflict}
 	  $resolve\_crash(crash)$\;
     }
     
	\eIf{$check\_finished()$}{\label{alg:solver_check_solved}
		$solved \leftarrow true$\;
	}{
		$resolve\_priority\_reset()$\; \label{alg:solver_priority_reset}
		$resolve\_replan()$\; \label{alg:solver_resolve_replan}
		$state\_shift(1)$\;\label{alg:solver_state_shift}
	} 
 }
 \caption{Robot maneuvering algorithm}
 \label{alg:solver}
\end{algorithm}

\subsubsection{Robot maneuvering phase}
This phase is described in Algorithm \ref{alg:solver}.
The loop (line \ref{alg:solver_solved}) is repeated until no conflict exists. 
First, it is checked whether there is a conflict in the current state of robots (line \ref{alg:solver_conflict}). 
Detected conflicts are immediately resolved by the {\em resolve\_crash} Algorithm. 
After that, it is checked whether all robots reached their final destination (line \ref{alg:solver_check_solved}). 
When the problem is not solved yet, temporal priorities are reset, and paths of robots blocked by already finished robots are replanned (lines~\ref{alg:solver_priority_reset}--\ref{alg:solver_resolve_replan}). 
Moreover, the time is incremented, and all robots are moved accordingly (line~\ref{alg:solver_state_shift}).

The \textit{conflict\_detect} procedure searches through all the blocked nodes and edges and checks whether they are occupied by some robot other than the currently tested one.

The \textit{resolve\_crash} algorithm solves one crash at the time. 
If several conflicts occur at the same time, only the first one is resolved; however, all the operations will cause the time to move one back; thus the other conflicts will also be resolved. 
The algorithm consists of two steps: it is first decided which operation from the set $\left\{ stop, push, replan\right\}$ will be used, and this operation is executed. 

\begin{figure}[!htb]
    \centering
    \begin{subfigure}[t]{0.48\columnwidth}
        \centering
        \includegraphics[width=0.6\textwidth]{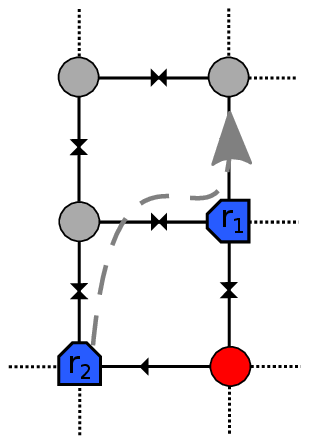}
        \caption{If $r_1$ is stopped, the robot $r_2$ would still crash into it.}
        \label{fig:stop_not_possible}
    \end{subfigure}\hfill
    \begin{subfigure}[t]{0.48\columnwidth}
        \centering
        \includegraphics[width=0.6\textwidth]{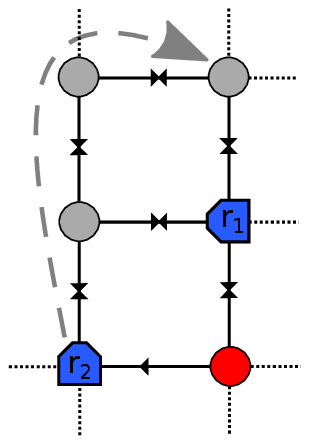}
        \caption{\opstop{} operation is possible. Stopping robot $r_1$ allows robot $r_2$ to move towards its goal.}
        \label{fig:stop_possible}
    \end{subfigure}
    \caption{Examples showing situations when the \opstop{} operation is possible and when it is not.}
\label{fig:stop_possibility}
\end{figure}

The decision is done in several steps.
As we do not allow to move a robot which is already in its final destination, replanning of the other robot in the conflict is invoked by the \opreplan{} operation. 
When no robot is finished, it is tested whether \opstop{} can be used by checking two conditions for both robots.
The first one is that the other robot's path does not cross the node that the tested robot would be stopped at. 
In Fig.~\ref{fig:stop_not_possible}, the robot $r_2$ has a path planned in a way, that stopping the robot $r_1$ would not help to resolve the conflict. If the path was planned differently (Fig.~\ref{fig:stop_possible}), the \opstop{} operation helps to resolve the conflict completely.

The other ond condition is implemented to prevent deadlock situations that might arise from \opstop{}. 
When the robot is stopped, it is put into the {\tt idle} state and waits until the first edge on its path is empty. 
To avoid possible deadlocks, the robot that is supposed to continue without stopping cannot have any idle robots on its way. 
When no other operation is selected, the \oppush{} operation is chosen.

\subsection{Operations}
The \opstop{} operation (Algorithm \ref{alg:stop}) aims to stop one robot so that the other one can continue without disruption. 
It is decided which robot should be stopped (line \ref{alg:stop_decide}) first. 
The algorithm already gets a set of one or two robots, which can be stopped and selects the one with a lower temporary priority. 
The time is then shifted back until the $r_{stop}$ is not occupying the conflicting node (line \ref{alg:stop_shift}).
$r_{stop}$ is thus in some other node $n_{s}$, where it  will be stopped. 

The wait sequence is generated (line \ref{alg:stop_generate_wait}) as a sequence of $steps\_shift$ time steps where the robot $r_{stop}$ stands still at the node $n_{s}$ ending with the time step with a special $idle$ flag. 
If this special time step is the next step, it is checked during the state shift (Algorithm \ref{alg:solver}, line \ref{alg:solver_state_shift}) whether the edge that the robot $r_{stop}$ is going to move at is not in conflict with any other robot. 
The wait sequence is extended to the moment there is no conflict. 
The generated wait sequence is then added to the path after the current state extending the planned trajectory $J_{r_{stop}}$ (line \ref{alg:stop_push_wait}).

\begin{algorithm}[!h]
\DontPrintSemicolon
 \KwData{Robots that crashed $r_1$ and $r_2$, robot model $L$}
 \KwResult{Stops one of the robots and resolves the conflict.}
$[r_{stop}, r_{go}] \leftarrow$ Decide which robot to stop\; \label{alg:stop_decide}
$steps\_shift$ $\leftarrow$ Shift time back until $r_{stop}$ is not occupying the conflicting node \label{alg:stop_shift}\;
	$J_{wait} \leftarrow$ Generate wait sequence using $L$.\;\label{alg:stop_generate_wait}
	Update $J_{r_{stop}}$ with $J_{wait}$\;\label{alg:stop_push_wait}
 \caption{\opstop{} operation}
 \label{alg:stop}
\end{algorithm}

\begin{figure}
    \centering
    \begin{subfigure}[b]{0.24\textwidth}
        \includegraphics[width=\textwidth]{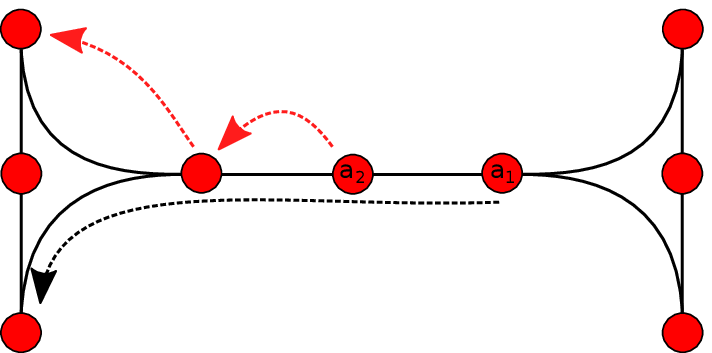}
        \caption{Original \oppush{} operation.}
        \label{fig:push_original}
    \end{subfigure}
    \hfill
    \begin{subfigure}[b]{0.24\textwidth}
        \includegraphics[width=\textwidth]{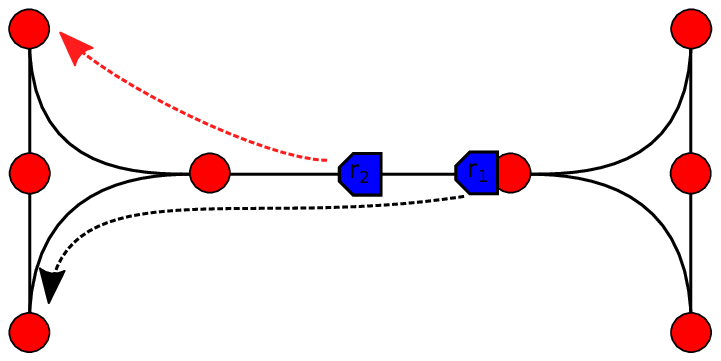}
        \caption{New \oppush{} operation.}
        \label{fig:push_new}
     \end{subfigure}
    \caption{Comparison of the original and new \oppush{} operation on simple case.}
\label{fig:push_ex}
\end{figure}

\begin{figure}
    \begin{subfigure}[b]{0.4\columnwidth}    
        \includegraphics[width=\textwidth]{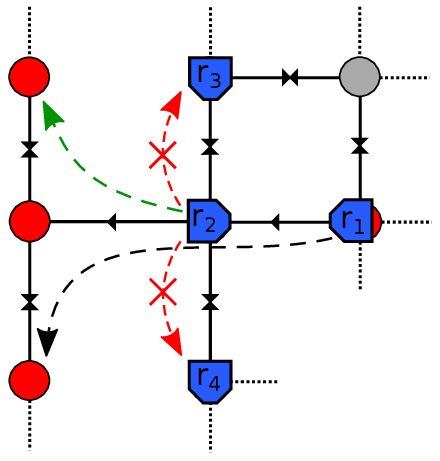}
        \caption{The robot $r_2$ cannot be pushed to the closest nodes, because those are occupied by finished robots $r_3$ and $r_4$ and must be pushed to the top left red node.}
        \label{fig:push_finished}
    \end{subfigure}
    \hfill
    \begin{subfigure}[b]{0.55\columnwidth}
        \raisebox{8.8mm}{        
        \includegraphics[width=\textwidth]{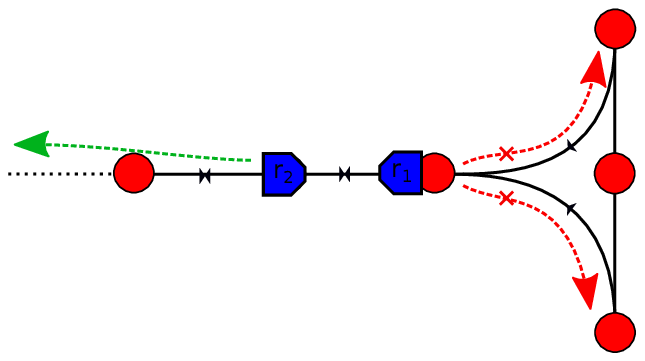}
        }
        \caption{It is impossible to push the robot $r_1$, because the only exiting edge ends on a node occupied by the robot $r_2$, which is trying to push it. The robot $r_2$ must be pushed.}
		\label{fig:push_impossible}
    \end{subfigure}
    \caption{Admissibility of the \oppush{} operation.}
\end{figure}

The \oppush{} operation is based on the same operation from the P\&R algorithm, which is used for the movement of agents even when there is no conflict. 
In the proposed algorithm, only the part of the operation that is invoked when the next node is occupied by an agent is considered, because it is used for resolving conflicts only and not moving the robots itself. 
The original \oppush{} moves a robot only by one node; thus the operation can be executed several times for a single robot if the robots have a conflict on a long isthmus as is illustrated in Fig.~\ref{fig:push_original} with the red arrows.
The new version moves the agents arbitrarily far when moving on an isthmus or when the closest nodes cannot be used for example if they are occupied by finished robots (Fig.~\ref{fig:push_finished}).

The new \oppush{} (Algorithm \ref{alg:push}) decides first which robot $r_{push}$ will be pushed away and which robot $r_{go}$ will continue on its path (line \ref{alg:push_decide}). 
The robot to be pushed is the one with a lower temporary priority, as it is less likely that the robot was pushed recently and the higher-priority robot has more likely a longer trajectory to travel through. 
In some situations, the chosen robots cannot be pushed due to a direction of adjacent edges, see, e.g., Fig.~\ref{fig:push_impossible}.
The other robot is selected in this case.

Next, \oppush{} finds the closest node to $r_{push}$ that is not on the path of $r_{go}$. 
List of nodes and edges that are on a path of $r_{go}$ are thus created (lines \ref{alg:push_blocknodes} and \ref{alg:push_blockedges}). 
Moreover, a list of nodes that are forbidden to expand during the search is determined. 
First, the node where  $r_{go}$ is going to wait is added to the node list to prevent robot $r_{push}$ from being pushed through this node and all nodes with finished robots are added since it is not possible to move them (line \ref{alg:push_no_exp_fin}). 
The path is then found (line \ref{alg:push_path}) using a modified version of Dijkstra's algorithm \cite{dijkstra} that accounts for the blocked edges and nodes with forbidden expansion.\;
The state of the algorithm is shifted back until the robot $r_{push}$ does not occupy a node (line \ref{alg:push_shift}) and all its future time steps are removed from its planned trajectory to be later replaced with the push trajectory (line \ref{alg:push_remove}).

The trajectory is generated using the robot model, and the $reset$ flag is added to the last generated step (line \ref{alg:push_traj}). 
The temporary priority of $r_{push}$ is increased by $r_{go}$ (line \ref{alg:push_prio}) to push other robots that might get into conflict with $r_{push}$ while being pushed. 
This way only a robot with a very high priority would be able to push this robot back. 
Priorities ensure that the non-finished robot with the highest priority always moves towards its destination. 
When the push is finished, the robot $r_{push}$ resets its temporary priority when the time step with the $reset$ flag is encountered in the Algorithm \ref{alg:solver} (line \ref{alg:solver_priority_reset}).

As the push of $r_{push}$ is executed on directed edges, moving back to the original position using the same path might be impossible. 
Instead, the algorithm generates a trajectory from the last node of $push\_path$ to the goal node of $r_{push}$ (line \ref{alg:push_replan}) using \opreplan{}. 
The trajectories are added together to form a new trajectory of the robot $r_{push}$.\;

The algorithm stops $r_{go}$ at the last visited node before the conflict. 
First, we need to shift the time to a state where $r_{go}$ is at this node. 
However, the state was shifted back before thus the state must be shifted by $(steps\_shift-t_{back\_to\_node})$, where $t_{back\_to\_node}$ is the number of steps from robot $r_{go}$ leaving the previous node before the state shift back (line \ref{alg:push_shift_go}). 
This shift can be either forward or backward. 
Similarly to \opstop{}, the operation generates the wait sequence $J_{wait}$ for $r_{go}$ with a minimal wait of $(steps\_shift-t_{back\_to\_node})$ steps to ensure that $r_{push}$ will get to the state of conflict. 
The special $idle$ flag is added to the last step of the wait (line \ref{alg:push_generate_wait}). 
The operation adds the $J_{wait}$ sequence into the $J_{r_{go}}$ trajectory right after the current step (line \ref{alg:push_update}) and the operation is complete.

\begin{algorithm}[ht]
\DontPrintSemicolon
 \KwData{Robots in conflict, robot model $L$, graph $G$}
 \KwResult{Resolves conflict with \oppush{} operation.}
$[r_{push}, r_{go}] \leftarrow$ Decide which robot to let go and which robot to push.\; \label{alg:push_decide}
$t_{back\_to\_node} \leftarrow$ Number of steps from when $r_{go}$ left last node.\;
$blocked\_nodes \leftarrow$ Nodes in path of $r_{go}$.\; \label{alg:push_blocknodes}
$blocked\_edges \leftarrow$ Edges in path of $r_{go}$.\; \label{alg:push_blockedges}
$no\_expansion\_nodes \leftarrow$ A node where $r_{go}$ waits or nodes with already finished robots.\; \label{alg:push_no_exp_fin}
$n_p \leftarrow$ Last node that $r_{push}$ occupied.\;
	$push\_path \leftarrow$ Find a path to closest node to $n_p$ with respect of $blocked\_nodes$, $blocked\_edges$ and $no\_expansion\_nodes$ on graph $G$.\; \label{alg:push_path}
	$[steps\_shift, t_{now}]$ $\leftarrow$ Shift state back until $r_{push}$ is occupying a node.\; \label{alg:push_shift}
	$J_{r_{push}} \leftarrow$ $J_{r_{push}}(j_0, \ldots , j_{(t_{now})})$.\; \label{alg:push_remove}
	$J_{push} \leftarrow$ Generate trajectory using $push\_path$ and $L$.\; \label{alg:push_traj}
	$J_{replanned} \leftarrow$ Generate trajectory from last node of $push\_path$ to goal node of $r_{push}$ using \opreplan{}.\; \label{alg:push_replan}
	$J_{r_{push}} \leftarrow$ $J_{r_{push}} \cup J_{push} \cup J_{replanned}$.\; \label{alg:push_final_traj}
	temporary priority of $r_{push} \leftarrow$ temporary priority of $r_{push} +$ temporary priority of $r_{go}$.\; \label{alg:push_prio}
	Shift state by $(steps\_shift-t_{back\_to\_node})$.\; \label{alg:push_shift_go}
	$J_{wait} \leftarrow$ Generate wait sequence with at least $(t_{back\_to\_node}-steps\_shift)$.\;\label{alg:push_generate_wait}
	Update $J_{r_{go}}$ with $J_{wait}$.\;\label{alg:push_update}
 \caption{The \oppush{} operation.}
 \label{alg:push}
\end{algorithm}

\begin{figure}
    \centering
        \includegraphics[width=0.3\columnwidth]{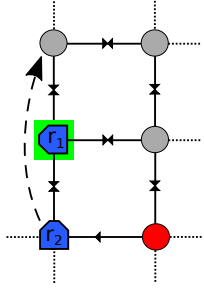}
        \caption{Both operations \opstop{} and \oppush{} would fail, because the robot $r_1$ is finished and cannot be moved. This situation requires the \opreplan{} operation.}
		\label{fig:push_skip}
\end{figure}

The \opreplan{} operation (Algorithm \ref{alg:replan}) is used when one of the robots in conflict is finished (Fig.~\ref{fig:push_skip}). 
The finished robots cannot be moved, thus \oppush{} and \opstop{} would not help in this case. 
The operation plans a new trajectory of the robot $r_x$ from its current node $n_c$ to the goal node $n_g$ while avoiding all nodes that the finished robots occupy. 
The property of a graph that by removing any storage location nodes, the graph will not become disconnected, is considered. 

At first, the algorithm identifies which of the robots is not finished (line \ref{alg:replan_choose}) to select the one that needs to be replanned. 
All nodes occupied by finished robots are added to the list $nodes\_to\_avoid$ (line \ref{alg:replan_avoid}) to ensure that they are avoided. 
The solution state is shifted back until the robot $r_{x}$ is not occupying the conflicting node (line \ref{alg:replan_shift}). 
The operation removes the planned trajectory of the robot $r_{x}$ from the current state to replace it further with the replanned one (line \ref{alg:replan_remove}). 
The algorithm then calculates the shortest path from the currently occupied node $n_c$ to the goal node $n_g$ using A* algorithm (line \ref{alg:replan_path}) assuming that all the nodes from $nodes\_to\_avoid$ are set as unreachable. 
A trajectory for $r_{x}$ describing its movement from node $n_{c}$ to its goal node $n_{g}$ is generated (line \ref{alg:replan_traj}) and added to the planned trajectory $J_{r_{x}}$ (line \ref{alg:replan_traj_add}).

Note that \opreplan{} operation removed a part of the planned trajectory and replaced it with a new one. 
If $r_{x}$ was performing the \oppush{} operation, the \textit{reset} flag for resetting the priority would be deleted. 
The algorithm resets the priority (line \ref{alg:replan_reset}) to avoid this loss because the change of the trajectory causes the robot is no longer performing the \oppush{} operation.

\begin{algorithm}[!h]
\DontPrintSemicolon
 \KwData{Robots in conflict, robot model $L$, graph $G$}
  \KwResult{A new trajectory $J_{r_{x}}$ of a non-finished robot $r_{x}$.}
    $r_{x} \leftarrow$ The robot that is not finished.\; \label{alg:replan_choose}
    $nodes\_to\_avoid \leftarrow$ Nodes occupied by finished robots.\; \label{alg:replan_avoid}
    $t_{now} \leftarrow$ Shift state back until $r_{x}$ is not occupying the current node.\; \label{alg:replan_shift}
    $n_c \leftarrow$ Node occupied by $r_{x}$.\;
    $n_g \leftarrow$ Goal node of $r_{x}$.\;
    $J_{r_{x}} \leftarrow$ $J_{r_{x}}(j_0, \ldots , j_{(t_{now})})$.\; \label{alg:replan_remove}
    $P \leftarrow $ Calculate the shortest path from $n_c$ to $n_g$ avoiding nodes in $nodes\_to\_avoid$.\;\label{alg:replan_path}
    $J_{replan} \leftarrow$ Generate trajectory using $P$ and $L$.\; \label{alg:replan_traj}
    $J_{r_{x}} \leftarrow J_{r_{x}} \cup J_{replan}$\; \label{alg:replan_traj_add}
    temporary priority of $r_{x} \leftarrow$ main priority of $r_{x}$.\; \label{alg:replan_reset} 
    
 \caption{\opreplan{} operation}
 \label{alg:replan}
\end{algorithm}


\subsection{Algorithm limitations}
\label{sec:limitations}
As already mentioned, limitations of the proposed algorithm are caused by simplifications made and specialization on the real environment.
Specifically, the directed graph must be connected and biconnected at prospective goal nodes except maintenance nodes.

Another limitation is the maximal number of robots in a given map. 
This number should be theoretically the same as in the P\&R algorithm, i.e., $n-1$ where $n$ is the number of robots.
Because no robot can finish on road nodes, the number of robots is thus $n-n_r$, where $n_r$ is the number of road nodes.
In the warehouse used during development and testing of the algorithm, the rate of robots to nodes (ignoring pick-station, isthmuses leading to and from pick-stations and queue parts of graph) is approximately $0.52$.

\subsection{Algorithm advantages}
\label{sec:adv}
The calculation of trajectories for a high number of robots in a big warehouse is computationally demanding. 
Also, the goals for robots do not have to be known in advance, and some might be added later on. 
The proposed algorithm solves both of these issues as discusses below.

The algorithm moves the state forward in time, and only when any conflict occurs, it moves the state back in time. 
For one operation, the time the state is moved back is the maximal time of the conflicted robots moving from the last node on the edge. 
However, if the operation causes another conflict before the operation is finished (e.g., stopping one robot causes a new conflict with another robot), the state could be moved back again. 
The amount of back-shifting is not limited, but long shifts are highly improbable. 
The algorithm can start running, buffer a solution for some time and then the robots can start moving in real time with s low risk of the solution state moving behind the real state of the warehouse. 
Implementation of the safety stop of the system when the state of the robots gets close to the state of the solution should be, of course, implemented. 
The buffering time must be decided by numerous simulations, and the available computational power must also be considered. 
The discussion about practical values of the buffering time is made in Section~\ref{sec:experiments}.

The algorithm allows for tasks being added during the calculation. 
The state of the solution must be moved back to the time when the new robot is supposed to start moving and the robot is simply added with its shortest path to its destination. 
It might cause new conflicts in previously calculated trajectories, but for conflicts that it does not affect, there is no need for recalculation, while the trajectories of these robots are already collision-free. 
One issue that might occur is that due to the impact of the newly added robot, some robots will perform operations that are no longer needed. 
For example, the newly added robot $r_1$ affects another robot $r_2$ that in the previous calculation would get into conflict with the robot $r_3$.
In the previous iteration, the robot $r_2$ pushed the robot $r_3$, and this trajectory was added to its trajectory.
In the next iteration, the robot $r_1$ stops the robot $r_2$, and it will not get into conflict with the $r_3$, but while the robot $r_3$ has the trajectory of the operation already calculated, it will still perform it. 
This might lead to unnecessary movements and delays.

\section{Experiments}
\label{sec:experiments}
The goal of the experiments is to assess the usability of the proposed algorithm in practice. 
The experiments were performed on a computer with an Intel i7-4771 processor and 8GB RAM running Linux Mint.

\begin{figure*}
    \begin{subfigure}[t]{0.32\textwidth}    
        \includegraphics[width=\textwidth]{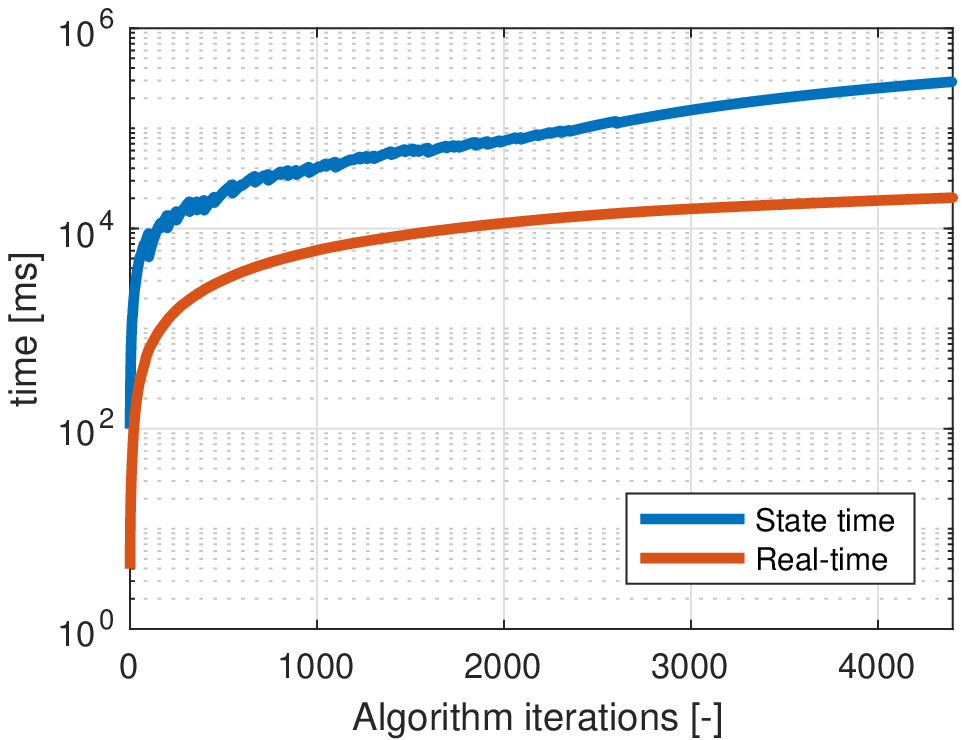}
        \caption{Real-time during the state time of the algorithm during the calculation of the \textit{sequential} approach.}
         \label{fig:time_exp}
    \end{subfigure}
    \hfill
    \begin{subfigure}[t]{0.32\textwidth}    
        \includegraphics[width=\textwidth]{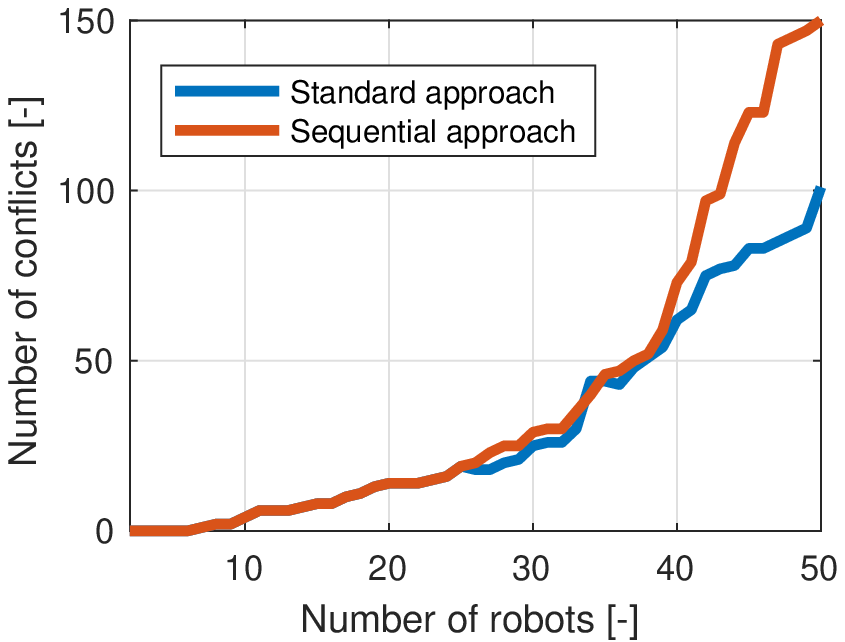}
        \caption{The number of conflicts for the \textit{sequential}  and \textit{standard} approaches.}
        \label{fig:crash_comp}
    \end{subfigure}
    \hfill
    \begin{subfigure}[t]{0.32\textwidth}    
        \includegraphics[width=\textwidth]{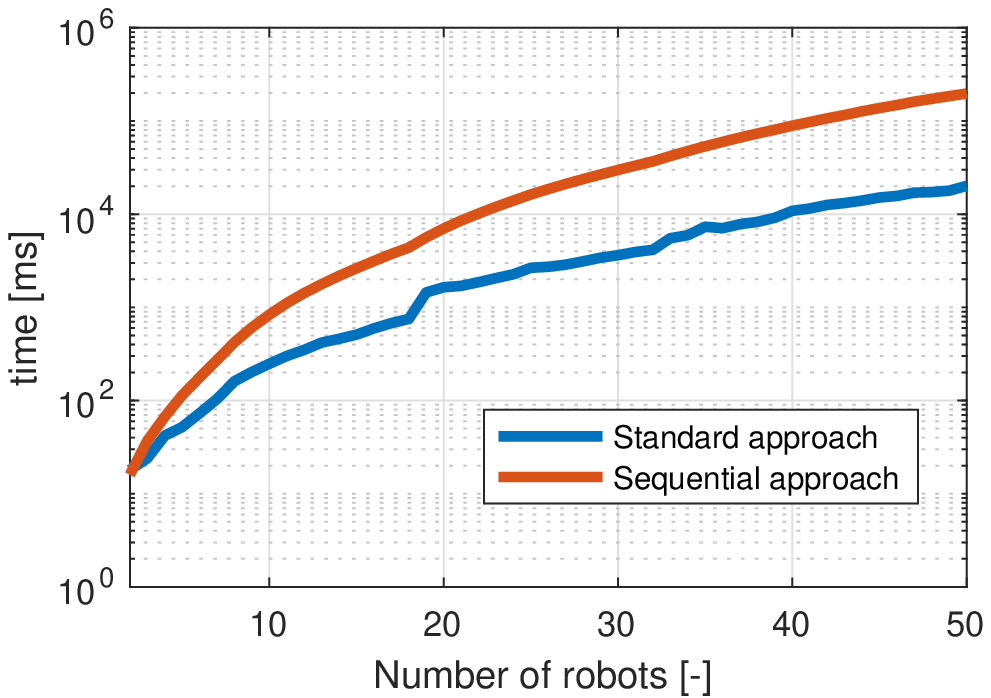}
        \caption{Calculation time of the \textit{sequential}  and \textit{standard} approaches..}
        \label{fig:calc_comp}
    \end{subfigure}
\end{figure*}

The warehouse map displayed in Fig.~\ref{fig:real_graph} was used for experiments. 
A task for $50$ robots (22 carrying a rack and 28 without it) with various distances between start and goal nodes was created randomly.  
Maintenance and storage location nodes were used as start and goal positions. 
The algorithm was executed with $2$ to $50$ robots to study the influence of the growing number of robots on the algorithm performance.

\subsection{Execution delay}
One of the main advantages of the algorithm is discussed in~Section \ref{sec:adv}: the algorithm moves forward in time, and calculated trajectories can be performed before the algorithm finishes. 
As the algorithm can move back in time, the risk of the execution being before the calculation must be addressed.

The algorithm was executed with $50$ robots and the time of the solution state was compared to real-time. 
In Fig.~\ref{fig:time_exp} one can see that the difference between state time and real-time grows (logarithmic scale was used for easier visual comparison); thus it is highly unlikely that the lines ever cross. 
In this case, the buffering time can be very small; thus the robots can start moving towards their goals instantly.

The more difficult the problem is (bigger warehouse, more robots), the higher is the risk of the execution catching up with the algorithm. 
This can be overcome with more computational resources and reasonable buffering time. 
One can also assume that not all robots will move at the same time. 
Some robots might be charging at the maintenance stations while others might be waiting in queue for a picking station.

\subsection{Two approaches comparison}
The ability of the algorithm to add robots to the plan during the calculation was also discussed in Section~\ref{sec:adv}. 
The extreme case of adding robots one by one to the beginning of the solution was tested to assess the impact on the results. 
The algorithm assumes that $k$ robots have already their plans and then a plan for the $k+1$th robot is calculated.  
This approach is named \textit{sequential}, while the approach of assuming all $n$ robots at once is named \textit{standard}. 

The number of conflicts to be solved for each amount of robots from $2$ to $50$ for both approaches was recorded and compared first. 
One can see in Fig.~\ref{fig:crash_comp} that the sum of conflicts for the \textit{sequential} approach is comparable with the \textit{standard} approach for the amount of robots ranging from $2$ to $39$. 
This shows that the newly added robots in this task only cause a few new conflicts with the comparison with the full calculation.
With the growing number of robots, however, the probability of long parts of the trajectories of robots being replanned due to the addition of new robots is growing.
The conflicts that have been solved previously might be thrown away with the trajectory, and thus new conflicts must be calculated. 
This result also confirms the expected property that the number of conflicts grows exponentially with the number of robots.

Perhaps the most significant impact of the \textit{standard} approach is on the solution time. 
Running the algorithm multiple times through the whole plan demands significantly more computational resources. 
In each run, fewer resources are needed since most conflicts were already solved. 
However, the cumulative value of calculation time for the \textit{sequential} approach is always significantly higher as seen in Fig.~\ref{fig:calc_comp} (a logarithmic scale is used for better visualization). 
In less extreme cases, adding a task during the calculation still causes a delay.
However, this is not as significant than the recalculation of the whole solution.

For example, $40$ robots start moving at the same time, and the algorithm gets far in front of the real execution. 
After a few seconds, when the algorithm is almost finished, a task for a robot is added $2$ seconds in advance to the real execution. 
The algorithm will go back and use the already calculated trajectories; thus only newly caused conflicts need to be calculated again. 
Some extreme cases might occur; thus it is very important to cautiously choose the optimal time reserve when adding a task for a robot.

\subsection{Solution quality}
\begin{figure}
    \centering
        \includegraphics[width=0.7\columnwidth]{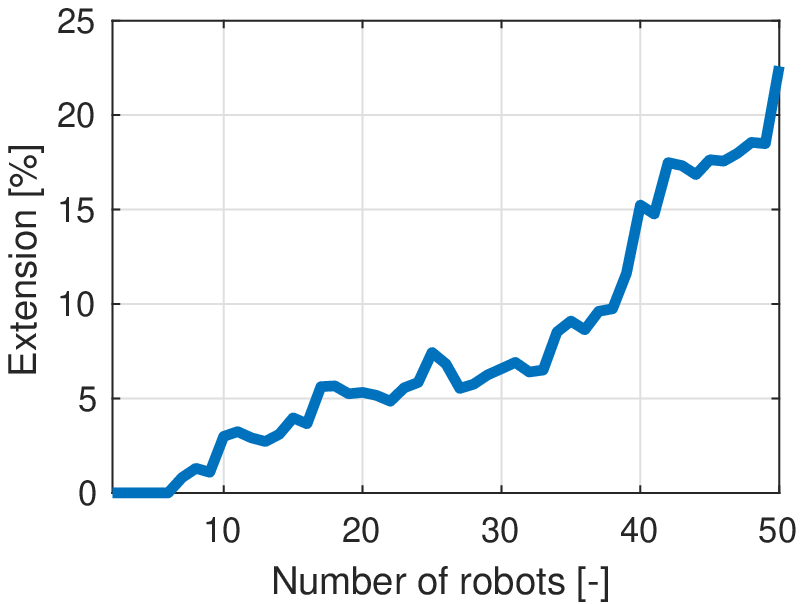}
        \caption{Extension of the sum of trajectories lengths  compared to the sum of the lengths of the lower-bound trajectories.}
		\label{fig:ext_comp}
\end{figure}

To measure the quality of generated trajectories, we compute a theoretical lower bound as a sum the shortest path length for each robot obtained by A* that ignores all
other robots in the assignment.
The ratio of the sum of trajectory steps for all robots provided by the proposed algorithms in comparison to this bound is shown in Fig.~\ref{fig:ext_comp}. 
This represents the effect of the operations on the trajectories with an increasing number of robots. 
There are no conflicts between the first $6$ robots; thus their cumulative trajectory length is the same as for the lowest threshold trajectories.
The cumulative trajectory length grows with the increasing number of conflicts. 
The figure is highly correlated with Fig.~\ref{fig:crash_comp} which represents the number of conflicts with the increasing number of robots. 

